\documentclass[conference]{IEEEtran}

\newcommand{\Fscore}{\ensuremath{\textit{F-score}}}%
\newcommand{\etal}{\textit{et al.}}
\usepackage{amsmath}
\usepackage{graphicx}
\usepackage{csquotes}%symbols
\usepackage{wasysym}%symbols
\usepackage[ruled]{algorithm2e}
\usepackage{multirow}
\usepackage{epstopdf}
\usepackage{amsmath}
\usepackage{amsfonts}
\usepackage{amssymb}
\usepackage{caption}
\usepackage{float}
\usepackage{subfig}
\usepackage[ruled]{algorithm2e}
\begin{document}

\title{NeCPD: An Online Tensor Decomposition with Optimal Stochastic Gradient Descent}

% use a multiple column layout for up to three different
% affiliations
\author{\IEEEauthorblockN{Ali Anaissi}
	\IEEEauthorblockA{School of Computer Science\\
		The University of Sydney\\
		Camperdown, NSW 2006, Australia\\
		Email: Ali.Anaissi@sydney.edu.au}
	\and
	\IEEEauthorblockN{Basem Suleiman}
	\IEEEauthorblockA{School of Computer Science\\
		The University of Sydney\\
		Camperdown, NSW 2006, Australia\\
		Email: basem.suleiman@sydney.edu.au}
		\and
	\IEEEauthorblockN{Seid Miad Zandavi}
	\IEEEauthorblockA{School of Computer Science\\
		The University of Sydney\\
		Camperdown, NSW 2006, Australia\\
		Email: miad.zandavi@sydney.edu.au}} 
\maketitle

% Copyright Statement
% When submitting your final paper to a SIAM proceedings, it is requested that you include 
% the appropriate copyright in the footer of the paper.  The copyright added should be 
% consistent with the copyright selected on the copyright form submitted with the paper.
% Please note that "20XX" should be changed to the year of the meeting.

% Default Copyright Statement

% Depending on which copyright you agree to when you sign the copyright form, the copyright 
% can be changed to one of the following after commenting out the default copyright statement
% above.

%\fancyfoot[R]{\scriptsize{Copyright \textcopyright\ 20XX\\
%Copyright for this paper is retained by authors}}

%\fancyfoot[R]{\scriptsize{Copyright \textcopyright\ 20XX\\
%Copyright retained by principal author's organization}}

%\pagenumbering{arabic}
%\setcounter{page}{1}%Leave this line commented out.

\begin{abstract}	
	
	Multi-way data analysis has become an essential tool for capturing underlying structures in higher-order datasets stored in  tensor  $\mathcal{X}  \in  \mathbb{R} ^{I_1 \times \dots   \times I_N} $.  $CANDECOMP/PARAFAC$  (CP) decomposition   has been extensively  studied and applied to approximate  $\mathcal{X}$ by $N$ loading matrices $A^{(1)}, \dots, A^{(N)}$ where $N$ represents the order of the tensor.     We propose a new efficient CP  decomposition solver  named NeCPD for non-convex problem in   multi-way online data based on stochastic gradient descent (SGD) algorithm. SGD  is very useful in online setting since it allows us to update $\mathcal{X}^{(t+1)}$ in one single step. In terms of global convergence, it is  well known that SGD  stuck  in many saddle points  when it deals with  non-convex problems. We study the Hessian matrix to identify theses saddle points,  and then try to escape them using  the perturbation approach which adds little noise to the gradient update step. We further apply    Nesterov's Accelerated Gradient (NAG)  method in SGD algorithm  to optimally accelerate the  convergence rate and compensate Hessian computational delay time per epoch. Experimental evaluation in the field of structural health monitoring using laboratory-based and real-life structural  datasets show that our method provides more accurate results compared with existing online tensor analysis methods.
	
\end{abstract}

\IEEEpeerreviewmaketitle

\section{Introduction}

The concept of multi-way data analysis was introduced by Tucker in 1964 as an extension of standard two-way data analysis to analyze multidimensional data known as tensor. It is often used when traditional two-way data analysis methods such as Non-negative Matrix Factorization (NMF), Principal Component Analysis (PCA) and Singular Value Decomposition (SVD) are not enough to capture the underlying structures inherited in multi-way data. In the realm of multi-way data, tensor decomposition methods such as $Tucker$ and $CANDECOMP/PARAFAC$ (CP) \cite{kolda2009tensor} have been extensively studied and applied in various fields including signal processing, civil engineer and time series analysis. The CP decomposition has gained much popularity for analyzing multi-way data due to its ease of interpretation. For example, given a tensor $\mathcal{X} \in \mathbb{R} ^{I_1 \times \dots \times I_N} $, CP method decomposes $\mathcal{X}$ by $N$ loading matrices $A^{(1)}, \dots, A^{(N)}$ each represents one mode explicitly, where $N$ is the tensor order and each matrix $A$ represents one mode explicitly. In contrast to  $Tucker$ method where the three modes can interact with each other making it difficult to interpret the resultant matrices. 

The CP decomposition often uses the Alternating Least Squares (ALS) method to find the solution for a given tensor. The ALS method follows the batch mode training process which iteratively solves each component matrix by fixing all the other components, then it repeats the procedure until it converges \cite{khoa2017smart}. However, ALS can lead to sensitive solutions  \cite{elden1980perturbation}\cite{anaissi2018regularized}. Moreover, in the realm of big and non-stationary data, the ALS method raises many challenges in dealing with data that is continuously measured at high velocity from different sources/locations and dynamically changing over time. For example,  structural health monitoring (SHM) data can be represented in a three-way form as  $location \times feature \times time$ which represents a large number of vibration responses measured over time by many sensors attached to a structure at different locations. This type of data can be found in many other application domains \cite{acar2009unsupervised,sun2008incremental,kolda2008scalable,anaissi2018tensor}. In such applications, a naive approach would be to recompute the CP decomposition from scratch for each new incoming  $X^{(t+1)}$. Therefore, this would become impractical and computationally expensive as new incoming datum would have a minimal effect on the current tensor.

Zhou et al. \cite{zhou2016accelerating} proposed a method called onlineCP to address the problem of online CP decomposition using ALS algorithm. The method was able to incrementally update the temporal mode in multi-way data but failed for non-temporal modes \cite{khoa2017smart}. In recent years, several studies have been proposed to solve the CP decomposition using stochastic gradient descent (SGD) algorithm which has the capability to deal with big data and online learning. However, such methods are inefficient and impractical due to slow convergence, numerical uncertainty and non-convergence \cite{ge2015escaping,anandkumar2016efficient,maehara2016expected,rendle2010pairwise}. 
To address the aforementioned problems, we propose an efficient solver method called NeCPD (Nesterov CP Decomposition) for analyzing  large-scale high-order data. The novelty of our proposed method is summarized in the following contributions:

\begin{enumerate}
	
	\item \textbf{Online CP Decomposition.} Our method is capable to update $\mathcal{X}^{(t+1)}$ in one single step. We realize this by employing the Stochastic Gradient Descent (SGD) algorithm to solve the CP decomposition as a baseline method.
	
	\item \textbf{Global convergence guarantee.} We followed the perturbation approach which adds a little noise to the gradient update step to reinforce the next update step to start moving away from a saddle point toward the correct direction. 
	
	\item \textbf{Optimal convergence rate.} Our method employs Nesterov's Accelerated Gradient (NAG) method into SGD algorithm to optimally accelerate the convergence rate \cite{sutskever2013importance}.  It achieves a global convergence rate of $O(\frac{1}{T^2})$ comparing to  $O(\frac{1}{T})$ for traditional SGD.
	
	\item \textbf{Empirical analysis on structural  datasets.} We conduct experimental analysis using laboratory-based and real-life  datasets in the  field of SHM. The experimental analysis shows that our method can achieve more stable and fast tensor decomposition compared to other known existing online and offline methods.      
\end{enumerate}

The remainder of this paper is organized as follows.  Section \ref{s:related} introduces background knowledge and review of the related work. Section \ref{s:method} describes our novel NeCPD algorithm for CP decomposition based on SGD algorithm augmented with NAG method and perturbation approach. Section \ref{s:motiv} presents the motivation of this work. Section \ref{s:results} shows the performance of NeCPD on structural datasets and presents our experimental results on both laboratory-based and real-life  datasets. The conclusion and discussion of future research work are presented in Section \ref{s:conclusion}.

\section{Background and Related work}
  \label{s:related}

\subsection{CP Decomposition}
Given a three-way tensor $\mathcal{X} \in \Re^{I \times J \times K} $,  CP  decomposes $\mathcal{X}$ into three  matrices  $A \in \Re^{I \times R}$, $B \in \Re^{J \times R} $and $ C \in \Re^{K \times R}$, where $R$ is the latent factors. It can be written as follows:

\begin{eqnarray}\label{eq:decomp}
\mathcal{X}_{(ijk)}  \approx \sum_{r=1}^{R}A_{ir} \circ B_{jr} \circ C_{kr}
\end{eqnarray}
where  "$\circ$" is a vector outer product.  $R$ is the latent element, $A_{ir}, B_{jr} $ and $C_{kr}$ are r-th columns of component 
matrices $A \in \Re^{I \times R}$, $B \in \Re^{J \times R} $and $ C \in \Re^{K \times R}$.
The main goal of  CP decomposition is   to decrease the sum  square error  between  the model and a given tensor $\mathcal{X}$. Equation \ref{eq:als} shows our loss function $L$ needs to be optimized.
 
\begin{eqnarray}\label{eq:als}	
L (\mathcal{X}, A, B, C) = \min_{A,B,C} \|  \mathcal{X} - \sum_{r=1}^R  \ A_{ir} \circ B_{jr} \circ C_{kr} \|^2_f,
\end{eqnarray}
where $\|\mathcal{X}\|^2_f$  is the  sum squares  of $\mathcal{X}$  and  the subscript $f$ is  the Frobenius norm. The loss function $L$ presented in Equation \ref{eq:als} is a non-convex problem with many local minima since it aims to optimize the sum squares of  three  matrices.  Several algorithms have been proposed to solve CP  decomposition \cite{symeonidis2008tag,lebedev2014speeding,rendle2009learning}. Among these algorithms, ALS  has been heavily employed which repeatedly solves each component matrix  by locking all other components until it   converges \cite{papalexakis2017tensors}. The rational idea of the least square algorithm is to set the partial  derivative of the loss function to zero with respect  to the parameter we need to minimize. Algorithm \ref{ALS} presents the detailed steps of ALS.

\begin{algorithm}
	%\renewcommand\thetable{Algorithm}
	%	\centering
	%
	%\label{ALS}
	\textbf{Alternating Least Squares\\}
	\textbf{Input}: Tensor $\mathcal{X} \in \Re^{I \times J \times K}  $, number of components $R$\\
	\textbf{Output}: Matrices  $A \in \Re^{I \times R}$, $B \in \Re^{J \times R} $ and  $ C \in \Re^{K \times R}$
	\begin{enumerate}
		\item[1:] Initialize $A,B,C$
		\item[2:] Repeat
		%	\begin{enumerate}
		{\setlength\itemindent{6pt}
			\item[3:] $A = \underset{A}{\arg\min} \frac{1}{2} \| X_{(1)} - A ( C \odot B)^T\|^2 $
			\item[4:] $B = \underset{B}{\arg\min} \frac{1}{2} \| X_{(2)} - B ( C \odot A)^T\|^2 $
			\item[5:] $C = \underset{C}{\arg\min} \frac{1}{2} \| X_{(3)} - C ( B \odot A)^T\|^2 $
			\item[]($X_{(i)} $ is the  unfolded matrix of $X$ in a current mode)	
		}
		%	\end{enumerate}
		\item[6:] until convergence
	\end{enumerate}
	\caption{  Alternating Least Squares for CP}
 	\label{ALS}
\end{algorithm}

\subsection{Stochastic Gradient Descent}
Stochastic gradient descent algorithm is a key tool for optimization problems. Lets say  our aim is to optimize a loss function  $L(x,w)$, where $x$ is a data point  drawn from a  distribution $\mathcal{D}$ and $w$ is a variable. The stochastic optimization problem can be defined as follows:
\begin{eqnarray}\label{eq:sgd}
w =  \underset{w}{argmin} \;    \mathbb{E}[L(x,w)]
\end{eqnarray}
The stochastic gradient descent method solves the above problem defined in Equation \ref{eq:sgd} by  repeatedly updates $w$ to minimize $L(x,w)$. It starts with some initial value of $w^{(t)}$ and then repeatedly performs the update as follows:
\begin{eqnarray}\label{eq:sgdu}
w^{(t+1)} :=    w^{(t)}   + \eta   \frac{\partial L}{\partial w } (x^{(t)} ,w^{(t)} )
\end{eqnarray}
where $\eta$ is the learning rate and $x^{(t)}$ is a random sample drawn from the given distribution $\mathcal{D}$.
This method guarantees the convergence of the loss function $L$ to the global minimum when it is convex. However, it can be susceptible to many local minima and saddle points when the loss function exists in a non-convex setting. Thus it becomes an NP-hard problem. In fact, the main bottleneck here is due to the existence of many saddle points and not to the local minima \cite{ge2015escaping}. This is because the rational idea of gradient algorithm depends only on the gradient information which may have $\frac{\partial L}{\partial u } = 0$ even though it is not at a minimum. 

Previous studies have used SGD for matrix factorization and tensor decomposition with extensions to handle the aforementioned issues.   Naiyang \etal \cite{guan2012nenmf}  applies  Nesterov's optimal gradient method to SGD for non-negative matrix factorization. This method accelerates the NMF process with less computational time. Similarly, Shuxin \etal \cite{zheng2017asynchronous} used an SGD algorithm for matrix factorization using Taylor expansion and Hessian information. They proposed a new asynchronous SGD algorithm to compensate for the delay resultant from a Hessian computation. 

Recently, SGD has attracted several researchers working on tensor decomposition.   For instance,  Ge \etal \cite{ge2015escaping} proposed a perturbed SGD (PSGD) algorithm for orthogonal tensor optimization. They presented several theoretical analysis that ensures convergence; however, the method is not applicable to non-orthogonal tensor. They also didn't address the problem of slow convergence. Similarly, Maehara \etal  \cite{maehara2016expected} propose a new algorithm for CP decomposition based on a combination of SGD and ALS methods (SALS). The authors claimed the algorithm works well in terms of accuracy. Yet its theoretical properties have not been completely proven and the saddle point problem was not addressed. Rendle and Thieme \cite{rendle2010pairwise} propose a pairwise interaction tensor factorization method based on  Bayesian personalized rank. The algorithm was designed to work only on three-way tensor data. To the best of our knowledge, this is the first work applies SGD algorithm augmented with Nesterov's optimal gradient and perturbation methods for optimal CP decomposition of multi-way tensor data.

\section{Nesterov CP Decomposition (NeCPD)}
\label{s:method}
Given an $N^{th}$-order tensor $\mathcal{X} \in \mathbb{R}^{I_1 \times \dots \times I_N}$, NeCPD  initially divides the CP problem into a convex $N$ sub-problems since its loss function $L$ is non-convex problem which may have many local minima. For simplicity, we present our method based on three-way tensor data. However, it can be naturally extended to handle a higher-order tensor. 

In a three-way tensor $\mathcal{X} \in \Re^{I \times J \times K} $,  $\mathcal{X}$ can be decomposed into three matrices $A \in \Re^{I \times R}$, $B \in \Re^{J \times R} $and $ C \in \Re^{K \times R}$, where $R$ is the latent factors. Following the SGD, we need to calculate the  partial  derivative of the loss function $L$ defined in Equation \ref{eq:als} with respect to the three modes $A, B$ and $C$ alternatively as follows:
\begin{eqnarray}\label{eq:partial}
\frac{\partial L}{\partial A }(X^{(1)}; A)  =   (X^{(1)} -   A \times  (C \circ B)) \times (C \circ B) \nonumber\\
\frac{\partial L}{\partial B }(X^{(2)}; B)  =   (X^{(2)} -   B  \times (C \circ A)) \times (C \circ A)\\
\frac{\partial L}{\partial C }(X^{(3)}; C)  =   (X^{(3)} -   C \times  (B \circ A)) \times (B \circ A)\nonumber
\end{eqnarray}
where $X^{(i)}$ is an unfolding matrix of tensor $\mathcal{X}$ in mode $i$. The gradient update step for $A, B$ and $C$ are as follows:
\begin{eqnarray}\label{eq:update}
A^{(t+1)} :=    A^{(t)}   + \eta^{(t)}   \frac{\partial L}{\partial A } (X^{(1, t)} ;A^{(t)} )  \nonumber\\
B^{(t+1)} :=    B^{(t)}   + \eta^{(t)}   \frac{\partial L}{\partial B } (X^{(2, t)} ;B^{(t)} )  \\
C^{(t+1)} :=    C^{(t)}   + \eta^{(t)}   \frac{\partial L}{\partial C } (X^{(3, t)} ;C^{(t)} )  \nonumber
\end{eqnarray}
The rational idea of SGD algorithm depends only on the gradient information of  $\frac{\partial L}{\partial w }$. In such non-convex setting, this  partial derivative may encounter data points with  $\frac{\partial L}{\partial w } = 0$ even though it is not at a global minimum. These data points are known as saddle points  which may detente the optimization process to reach the desired local  minimum if not escaped  \cite{ge2015escaping}. These saddle points  can be  identified by studying the second-order derivative  (aka Hessian)  $\frac{\partial L}{\partial w }^2$. Theoretically, when the $\frac{\partial L}{\partial w }^2(x;w)\succ  0$, $x$ must be a local minimum; if $\frac{\partial L}{\partial w }^2(x;w) \prec 0$, then we are at a local maximum; if $\frac{\partial L}{\partial w }^2(x;w)$ has both positive and negative eigenvalues, the point is a saddle point. The second order methods guarantee  convergence, but the  computing of Hessian matrix $H^{(t)}$ is  high,  which makes the method infeasible for high dimensional data and online learning. Ge \etal \cite{ge2015escaping} show that saddle points are very unstable and can be escaped if we slightly perturb them with some noise. Based on this, we use the perturbation approach which adds Gaussian noise to the gradient. This reinforces the next update step to start moving away from that saddle point toward the correct direction. After a random perturbation, it is highly unlikely that the point remains in the same band and hence it can be efficiently escaped (i.e., no longer a saddle point). Since we interested in fast optimization process due to online settings, we further incorporate Nesterov's method into the perturbed-SGD algorithm to accelerate the convergence rate. Recently, Nesterov's Accelerated Gradient (NAG) \cite{nesterov2013introductory} has received much attention for solving convex optimization problems \cite{guan2012nenmf,nitanda2014stochastic,ghadimi2016accelerated}. It introduces a smart variation of momentum that works slightly better than standard momentum. This technique modifies the traditional SGD by introducing velocity $\nu$ and friction $\gamma$, which tries to control the velocity and prevents overshooting the valley while allowing faster descent. Our idea behind Nesterov's is to calculate the gradient at a position that we know our momentum is about to take us instead of calculating the gradient at the current position. In practice, it performs a simple step of gradient descent to go from $w^{(t)} $ to $w^{(t+1)}$, and then it shifts slightly further than $w^{(t+1)}$ in the direction given by $\nu^{(t-1)}$. In this setting, we model the gradient update step with NAG as follows: 
\begin{eqnarray}\label{eq:nagNe}
A^{(t+1)} :=    A^{(t)}   + \eta^{(t)}   \nu^{(A, t)} +  \epsilon - \beta ||A||_{L_1} 
\end{eqnarray}
where
\begin{eqnarray}\label{eq:velNe}
\nu^{(A, t)} :=    \gamma \nu^{(A, t-1)}  + (1-\gamma)  \frac{\partial L}{\partial A } (X^{(1, t)} ,A^{(t)} ) 
\end{eqnarray}
where $\epsilon$ is a Gaussian noise, $\eta^{(t)}$ is the step size,  and $||A||_{L_1}$ is the regularization and penalization parameter into the $L_1$ norms to achieve smooth representations of the outcome and thus bypassing the perturbation surrounding the local minimum problem. The updates for $(B^{(t+1)} , \nu^{(B, t)})$ and $(C^{(t+1)}  ,\nu^{(C, t)} )$ are similar to the aforementioned ones.
With NAG, our method achieves a global convergence rate of $O(\frac{1}{T^2})$ comparing to $O(\frac{1}{T})$  for traditional gradient descent. Based on the above models, we present our NeCPD algorithm \ref{NeCPD}.
\begin{algorithm}
	%\renewcommand\thetable{Algorithm}
	%	\centering
	\caption{  NeCPD algorithm}
	\label{NeCPD}
\textbf{NeCPD}\\
	\textbf{Input}: Tensor $X \in \Re^{I \times J \times K}  $ , number of components $R$\\
	\textbf{Output}: Matrices  $A \in \Re^{I \times R}$, $B \in \Re^{J \times R} $ and  $ C \in \Re^{K \times R}$
	\begin{enumerate}
		\item[1:] Initialize $A,B,C$
		\item[2:] Repeat
		%	\begin{enumerate}
		{\setlength\itemindent{6pt}
			\item[3:] Compute the partial derivative of $A, B$ and $C$ using Equation \ref{eq:partial}
			\item[3:] Compute $\nu$ of $A, B$ and $C$ using Equation \ref{eq:velNe}
			\item[4:] Update the matrices  $A, B$ and $C$ using Equation \ref{eq:nagNe}	
		}
		%	\end{enumerate}
		\item[6:] until convergence
	\end{enumerate}
	
\end{algorithm}

\section{Motivation}
  \label{s:motiv}
Numerous types of data are naturally structured as multi-way data. For instance, structural health monitoring (SHM) data can be represented in a three-way form as  $location \times feature \times time$. Arranging and analyzing the SHM data in multidimensional form would allow to  capture the correlation between sensors  at different locations and  at the same time which was not possible using the standard two-way matrix $time\times feature$. Furthermore, in SHM only positive data instances i.e healthy state are available. Thus, the problem becomes an anomaly detection problem in higher-order datasets.  Rytter \cite{rytter1993vibrational} affirms that damage identification requires also damage localization and severity assessment which are considered much more complex than damage detection since they require a supervised learning approach \cite{worden2006application}. 

%To address the above problems in SHM applications, we employ our NeCPD method to learn from SHM data in multiple modes at the same time, and we use one-class SVM \cite{scholkopf2000support} as an anomaly detection method. The rationale behind one-class SVM is to map a positive data into a feature space using a kernel method. Recently, the Gaussian kernel has gained much popularity in many application domains. It has a parameter denoted $\sigma$ which may profoundly affect the performance of a one-class SVM by over-fitting or under-fitting the model. In our NeCPD approach, we use Edged Support Vector (ESV) algorithm \cite{anaissi2018gaussian} to tune $\sigma$ as it has the capability to work in a one-class learning setting.

Given a positive three-way SHM data $\mathcal{X} \in  \mathbb{R}^{feature \times location \times time}$, NeCPD decomposes $\mathcal{X}$ into three matrices $A, B$ and $C$. The $C$ matrix represents the temporal mode where each row contains information about the vibration responses related to an event at time $t$. The analysis of this component matrix can help to detect the damage of the monitored structure. Therefore, we use the $C$ matrix to build a one-class anomaly detection model using only the positive  training events. For each new incoming $\mathcal{X}_{new}$, we update the three matrices $A, B$ and $C$ incrementally as described in Algorithm \ref{NeCPD}. Then the constructed model estimates the agreement between the new event $C_{new}$ and the trained data. 

For damage localization,  we analyze the data in the location matrix $B$, where each row captures meaningful information for each sensor location. When the matrix $B$ is updated due to the arrival of a new event $\mathcal{X}_{new}$, we study the variation of the values in each row of matrix $B$ by computing the average distance from $B$'s row to $k$-nearest neighboring locations as an anomaly score for damage localization.
For severity assessment in damage identification, we study the decision values returned from the one-class  model. This is because a structure with more severe damage will behave much differently from a normal one.
\section{Experimental Results}
\label{s:results}
We conduct all our experiments  using an 	Intel(R) Core(TM) i7 CPU 3.60GHz with 16GB memory. We use R language to implement our algorithms  with the help of the two packages  \textbf{rTensor}
and \textbf{e1071} for tensor tools and one-class model.

\subsection{Comparison on synthetic data}
\begin{figure}[!t]
	\centering
	\includegraphics[scale=0.6]{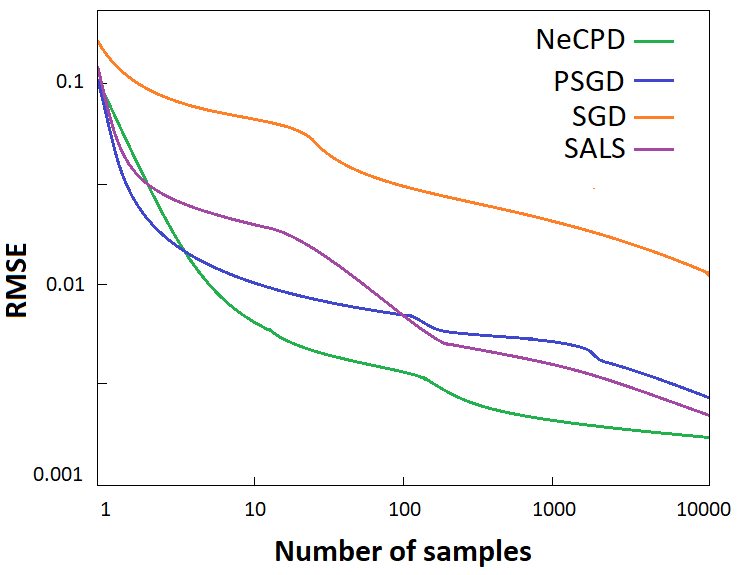}
	\caption{ Comparison of algorithms in terms of robustness and convergence. }
	\label{conv}
\end{figure}

Our initial experiment was to compare our NeCPD method to SGD, PSGD,  and SALS algorithms in terms of robustness and convergence. We generated a synthetic long time dimension   tensor $ \mathcal{X} \in \Re^{60 \times 12 \times 10000}$ from 12 random   loading matrices ${M}_{i=1}^{12} \in   \Re^{10000 \times 60}$ in which entries were drawn from uniform distribution $\mathcal{D} [0,1]$. We evaluated the performance of each method by plotting the number of sample $t$ versus the root mean square error (RMSE). For all experiments we use the learning rate $\eta^{(t)} =  \frac{1}{1 + t}$. It  can be clearly seen from Figure \ref{conv}  that  NeCPD outperformed the SGD and PSGD algorithms in terms of convergence and robustnesses.  Both SALS and NeCPD converged to a small RMSE but it was lower and faster in NeCPD.

\subsection{Comparison on SHM data}
We evaluate our NeCPD algorithm on laboratory-based and real-life structural datasets in the area of SHM. The two datasets are naturally in multi-way form. The first dataset is collected from a cable-stayed bridge in  Western Sydney, Australia. The second one is a specimen building structure obtained from Los Alamos National Laboratory (LANL) \cite{larson1987alamos}. 

\subsubsection{The Cable-Stayed Bridge:}
\label{s:data_wsu}

\begin{figure*}
	\centering
	\includegraphics[scale=0.75]{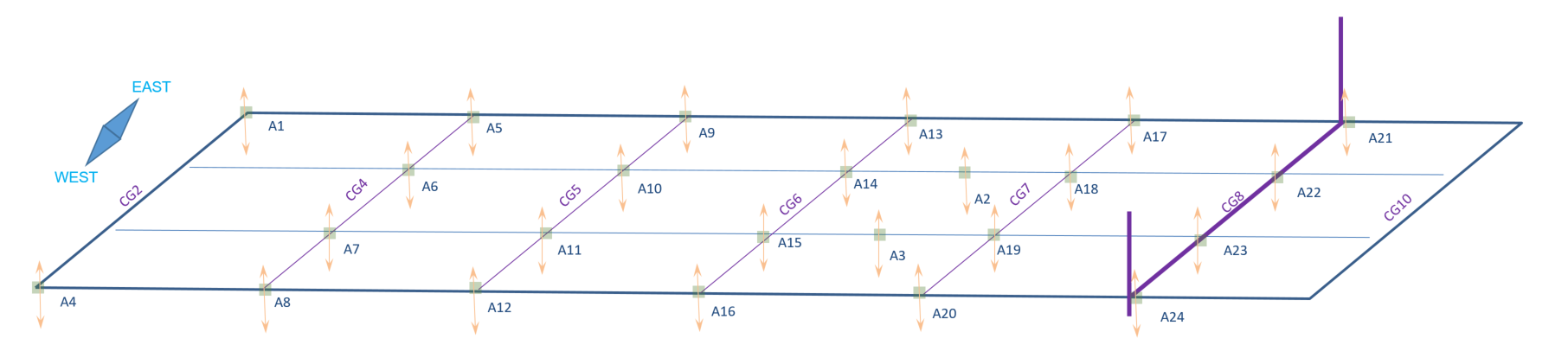}
	\caption{The locations on the bridge's deck of the 24 $Ai$ accelerometers used in this study. The cross girder $j$ of the bridge is displayed as $CGj$ \cite{anaissi2018tensor}.}
	\label{fig:wsuloc}
\end{figure*}

We used 24 uniaxial accelerometers and 28 strain gauges to measure the vibration and strain responses of the bridge at different locations. However, only accelerations data collected from sensors $Ai$ with $i\in [1;24]$ were used in this study.   Figure~\ref{fig:wsuloc} illustrates the locations of the 24 sensors on the bridge deck. Since the bridge is in healthy condition and in order to evaluate the performance of our method in damage identification tasks, we placed two different stationary vehicles with different masses to emulate two different levels of damage severity  \cite{kody2013identification,cerda2012indirect}.  Three different categories of data were collected  in this experiment: "Healthy-Data" when the bridge is free of vehicles;  "Car-Damage"   when a  light car vehicle is placed on the bridge close to location $A10$; and  "Bus-Damage" when a heavy bus vehicle is located on the bridge at location $A14$. This experiment generates 262 samples (i.e., events) separated into three categories: "Healthy-Data" (125 samples), "Car-Damage" data (107 samples) and "Bus-Damage" data(30 samples). Each event consists of acceleration data for a period of 2 seconds sampled at a  rate of 600Hz. The resultant event's feature vector composed of 1200 frequency values.

\subsubsection{Building Data:}
\label{s:data_b}
\begin{figure}
	\centering
	\includegraphics[scale=0.35]{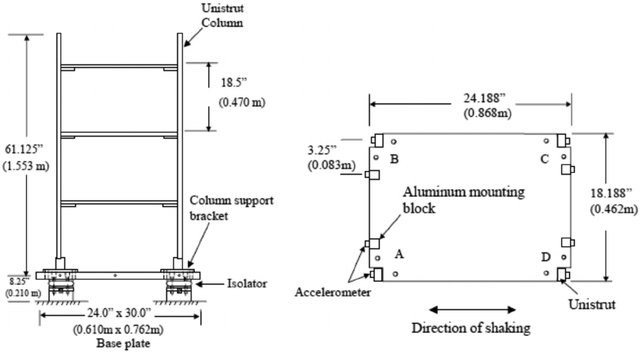}
	\caption{Three-story building and floor layout \cite{larson1987alamos}.}
	\label{fig:alamos}
\end{figure}

This data is based on experiments conducted by LANL \cite{larson1987alamos} using a specimen for a three-story building structure as shown in Figure \ref{fig:alamos}. Each joint in the building was instrumented by two accelerometers.  The excitation data was generated using a shaker placed at corner $D$. Similarly, for the sake of damage detection evaluation, the damage was simulated by detaching or loosening the bolts at the joints to induce the aluminum floor plate moving freely relative to the Unistrut column. Three different categories of data were collected in this experiment: "Healthy-Data" when all the bolts were firmly tightened;  "Damage-3C" data when the bolt at location 3C was loosened; and  "Damage-1A3C" data when the bolts at locations 1A and 3C were loosened simultaneously. This experiment generates 240 samples (i.e., events) which also were separated into three categories: Healthy-Data (150 samples), "Damage-3C" data (60 samples) and "Damage-1A3C" data(30 samples). The acceleration data was sampled at 1600 Hz. Each event was measured for a period of 5.12 seconds resulting in a vector of 8192 frequency values.
 
\subsubsection{Feature Extraction:}
\label{section:fe}
The raw signals of the sensing data we collected in the aforementioned experiments exist in the time domain. In practice, time domain-based features may not capture the physical meaning of the structure. Thus, it is noteworthy to convert the generated data to a frequency domain. For all datasets,  we initially normalize the time-domain features to have zero mean and one standard deviation. Then we use the fast Fourier transform method to convert them into the frequency domain. The resultant three-way data collected from the  Cable-Stayed Bridge now has a structure of  600 features  $\times$ 24 sensors  $\times$ 262 events.  For Building data,  we compute the difference between signals of two adjacent sensors which leads to 12 different joints in the three stories as in \cite{larson1987alamos}.  Then we select the first 150 frequencies as a feature vector resulting in a three-way data with a structure of 768 features $\times$ 12 locations  $\times$ 240 events.

\subsubsection{Experimental Setup:}
\label{section:setup}
For all case studies we apply the following procedures:
\begin{itemize}
	
	\item  The bootstrap technique to randomly select 80\% of the healthy samples for training and the remaining 20\% for testing in addition to the damage samples. All the reported accuracies are based on the average results over ten trials of the bootstrap experiment.

	\item The core consistency diagnostic (CORCONDIA) technique described in \cite{bro2003new} is used  to decide the number of rank-one tensors $\mathcal{X}$ in the NeCPD.
	
	\item    We use one-class support vector machine (OSVM) \cite{scholkopf2000support} as our one class model for anomaly detection method. The Gaussian kernel parameter  $\sigma$ in OCSVM is  tuned using the  Edged Support Vector (ESV) algorithm \cite{anaissi2018gaussian}, and the rate of anomalies $\nu$  was set to 0.05.

	\item  The $\Fscore$  measure to compute the accuracy values.   It is defined as $\textrm{\Fscore} = 2 \cdot \dfrac{\textrm{Precision}  \times \textrm{Recall} }{\textrm{Precision} + \textrm{Recall}}$ where $\textrm{Precision} = \dfrac{\textrm{TP} }{\textrm{TP} + \textrm{FP}}$ and $\textrm{Recall}  = \dfrac{\textrm{TP} }{\textrm{TP} + \textrm{FN}}$ (the number of true positive, false positive and false negative are abbreviated by TP, FP and FN, respectively). 
	
	\item The results of the competitive  method SALS proposed in \cite{maehara2016expected}  are compared against our NeCPD method.
	
\end{itemize}

\subsubsection{Results and Discussions:}
\label{chapter:results}
\subsubsection{The Cable-Stayed Bridge:}
\label{s:wsueval}
Our NeCPD method with one-class SVM was initially validated using vibration data collected from the cable-stayed bridge described in Section \ref{s:data_wsu}. The healthy training three-way tensor data (i.e. \textbf{training} set) was in the form of $ \mathcal{X} \in \Re^{24 \times 600 \times 100}$. The 137 examples related to the two damage cases were added to the remaining 20\% of the healthy data to form a \textbf{testing} set, which was later used for the model evaluation. This experiment generates a damage detection  accuracy $\Fscore$  of $1 \pm 0.00$ on the \textbf{testing} data. On the other hand, the $\Fscore$     accuracy of one-class SVM using SALS  is recorded at $0.98 \pm 0.02$.

\begin{figure}[!t]
	\centering
	\captionsetup[subfloat]{justification=centering}
	\subfloat[NeCPD.] {{\includegraphics[height=1.8in,width=1.7in]{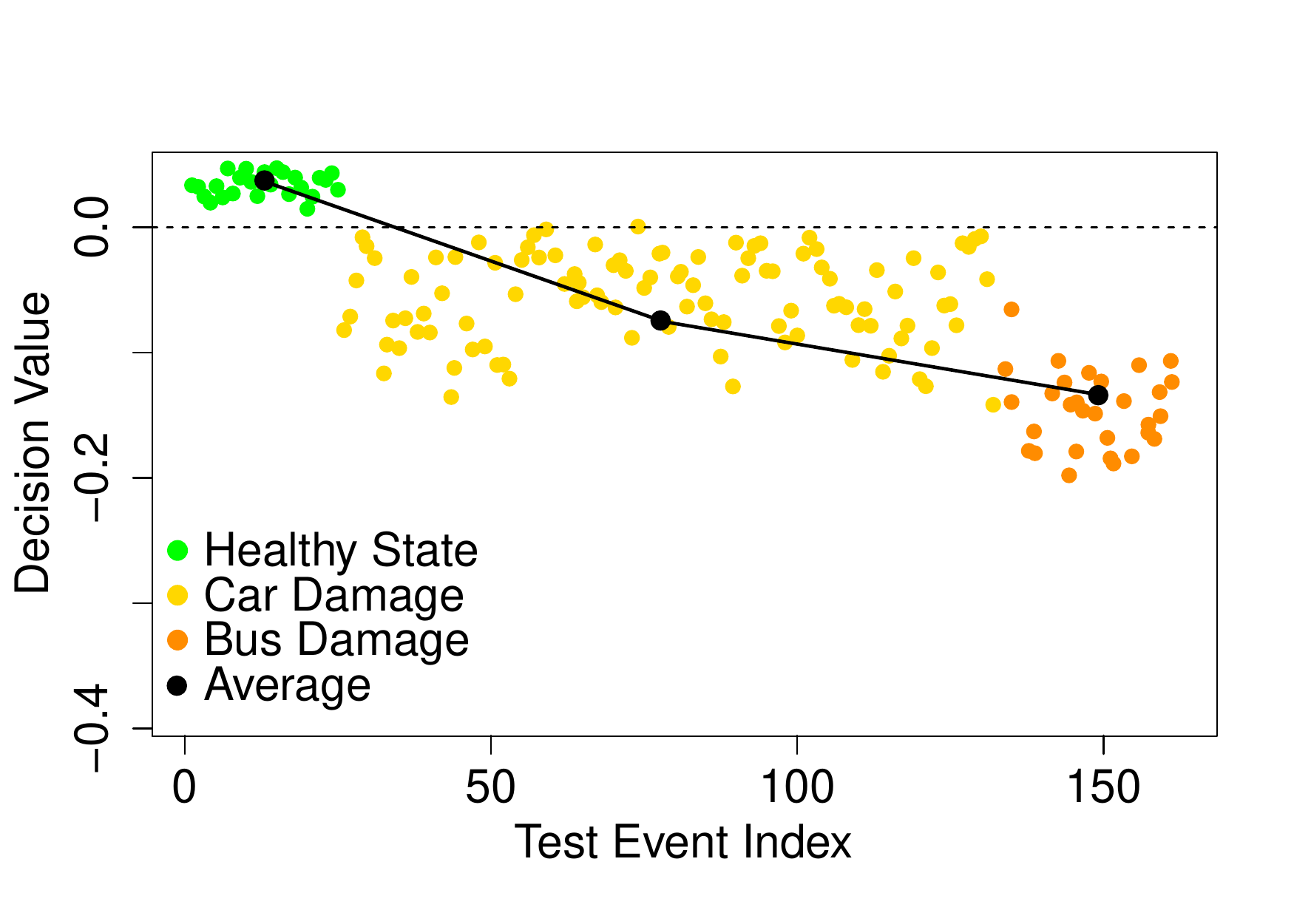} }}	
	\subfloat[ SALS.] 
	 {{\includegraphics[height=1.8in,width=1.7in]{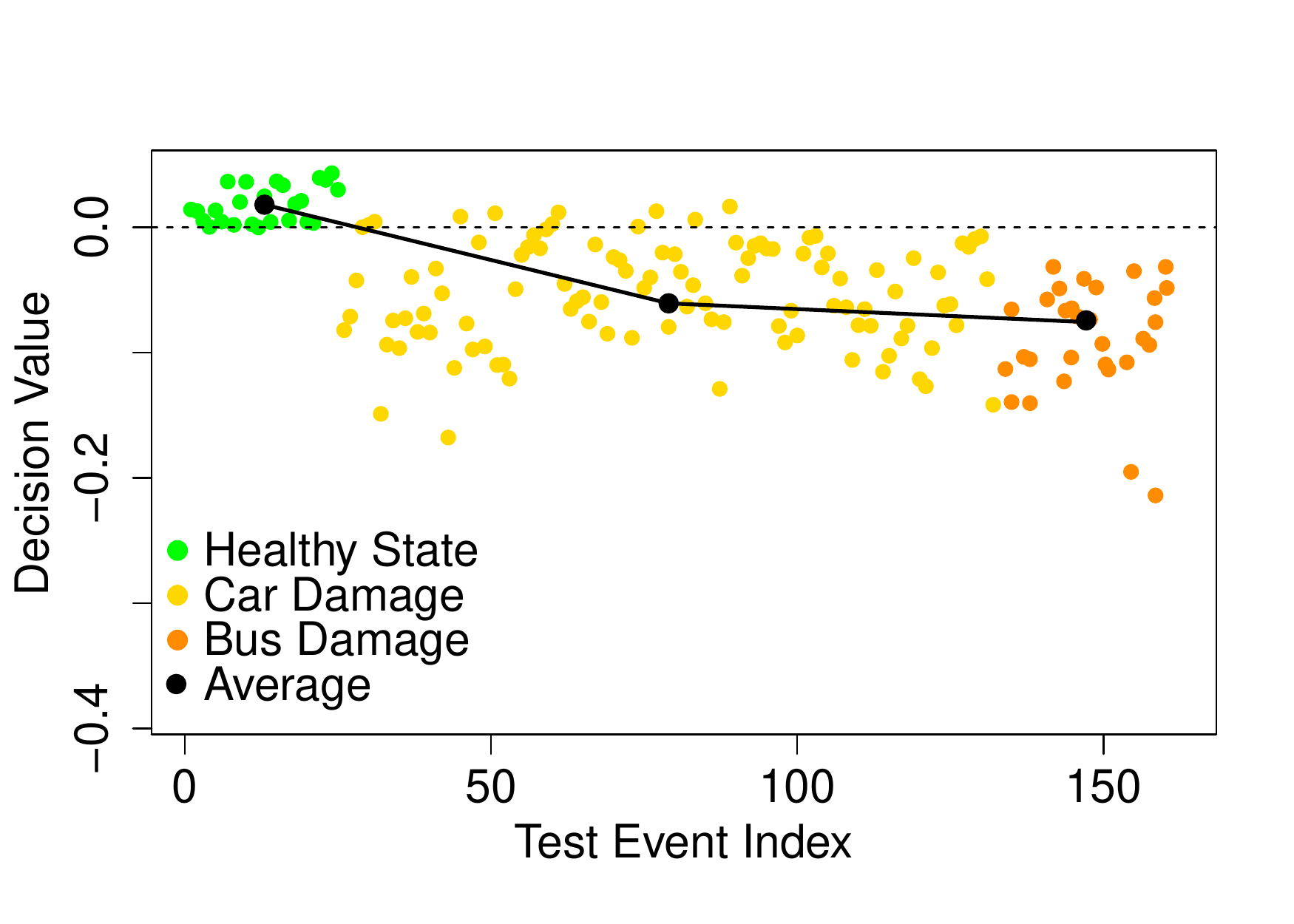} }}%
	\caption{Damage estimation applied on Bridge data using decision values obtained by  one-class SVM.}%
	\label{dv_est}
\end{figure}

As we anticipated, tensor analysis with NeCPD is able to capture the underlying structure in multi-way data with better convergence. This is demonstrated by plotting the decision values returned from one-class SVM based NeCPD (as shown in Figure \ref{dv_est}). We can clearly separate the two damage cases ("Car-Damage" and "Bus-Damage") in this dataset where the decision values are further decreased for the samples related to the more severe damage cases (i.e., "Bus-Damage"). These results suggest using the decision values obtained by NeCPD and one-class SVM as structural health scores to identify the damage severity in a one-class aspect. In contrast, the resultant decision values of one-class SVM based on SALS  are also able to track the progress of the damage severity in the structure but with a slight decreasing trend in decision values for "Bus-Damage" (see Figure \ref{dv_est}).

\begin{figure}[!t]
	\captionsetup[subfloat]{justification=centering}
	\subfloat[NeCPD.] {{\includegraphics[trim=0cm 1cm 3cm 0cm,clip=true,height=2in,width=3.5in]{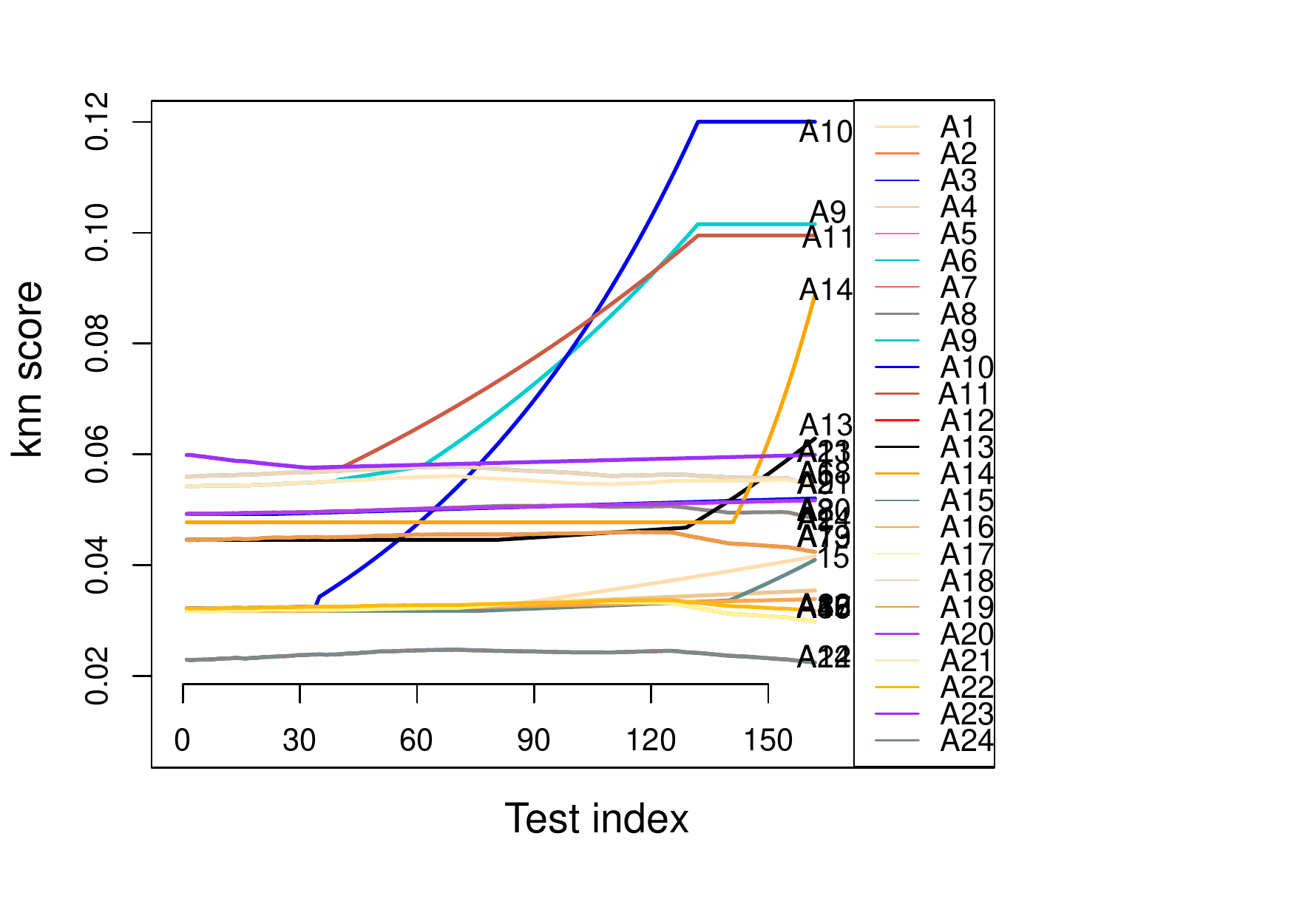} }}
	
	\subfloat[ SALS.] {{\includegraphics[trim=0cm 1cm 3cm 0cm,clip=true,height=2in,width=3.5in]{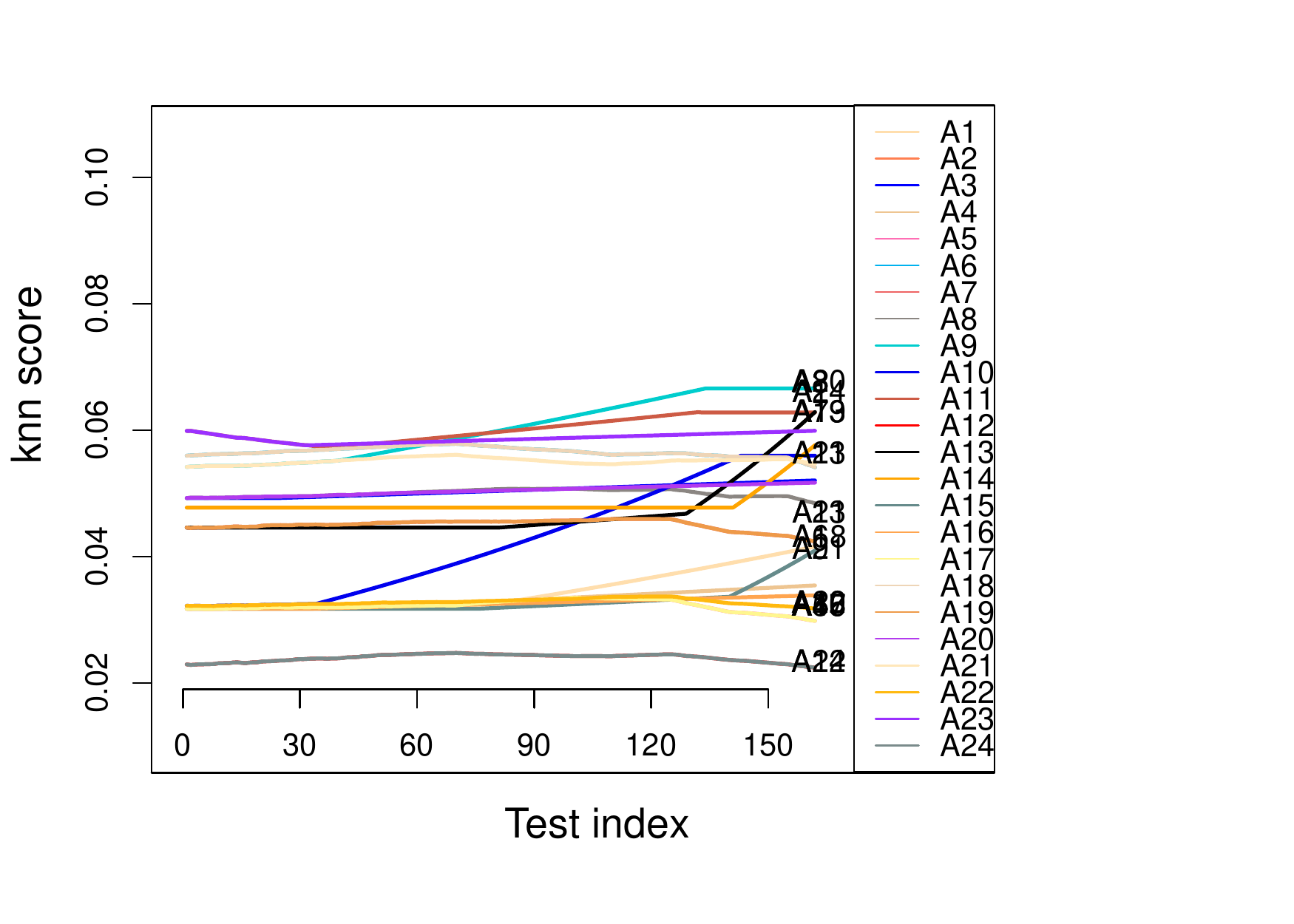} }}%
	
	%\subfloat[ batchCP-ALS.] {{\includegraphics[trim=0cm 1cm 3cm 0cm,clip=true,height=2in,width=3.5in]{bridge_batch_loc} }}%
	\caption{ Damage localization for the Bridge data: NeCPD  successfully localized damage locations.}
	\label{fig:wsu_necpd_loc}
\end{figure}

The last step was to analyze the location matrix $B$ obtained by NeCPD to locate the detected damage. Each row in this matrix captures meaningful information for each sensor location. Therefore,  we calculate the average distance from each row in the matrix $B_{new}$ to $k$-nearest neighboring rows. Figure \ref{fig:wsu_necpd_loc} shows the obtained $k$-nn score for each sensor. The first 25 events (depicted on the x-axis) represent healthy data, followed by 107 events related to "Car-Damage" and 30 events to "Bus-Damage". It can be clearly observed that   NeCPD method successfully localized the damage in the structure. Whereas,    sensors $A10$ and $A14$ related to the "Car-Damage" and "Bus-Damage" respectively behave significantly different from all the other sensors apart from the position of the introduced damage. Also, we observe that the adjacent sensors to the damage location (e.g $A9$, $A11$, $A13$ and $A15$) react differently due to the arrival pattern of the damage events. The SALS method, however, is not able to successfully locate the damage since it fails to update the location matrix $B$ incrementally.

\subsubsection{Building Data:}
 
Our second experiment is conducted using the acceleration data acquired from 24 sensors instrumented on the three-story building  as described in Section \ref{s:data_b}. The healthy training three-way data (i.e.\textbf{training} set) was in the form of  $ X \in \Re^{12 \times 768 \times 120}$. The remaining 20\% of the healthy data and the data obtained from the two damage cases were used for testing (i.e.\textbf{testing} set). The NeCPD with one-class SVM achieve an $\Fscore$ of $95 \pm 0.01$  on the \textbf{testing} data compared to $0.91 \pm 0.00$ obtained from one-class SVM based SALS. 

\begin{figure}[!t]
	\centering
	\captionsetup[subfloat]{justification=centering}
	\subfloat[NeCPD.] {{\includegraphics[height=1.5in,width=1.7in]{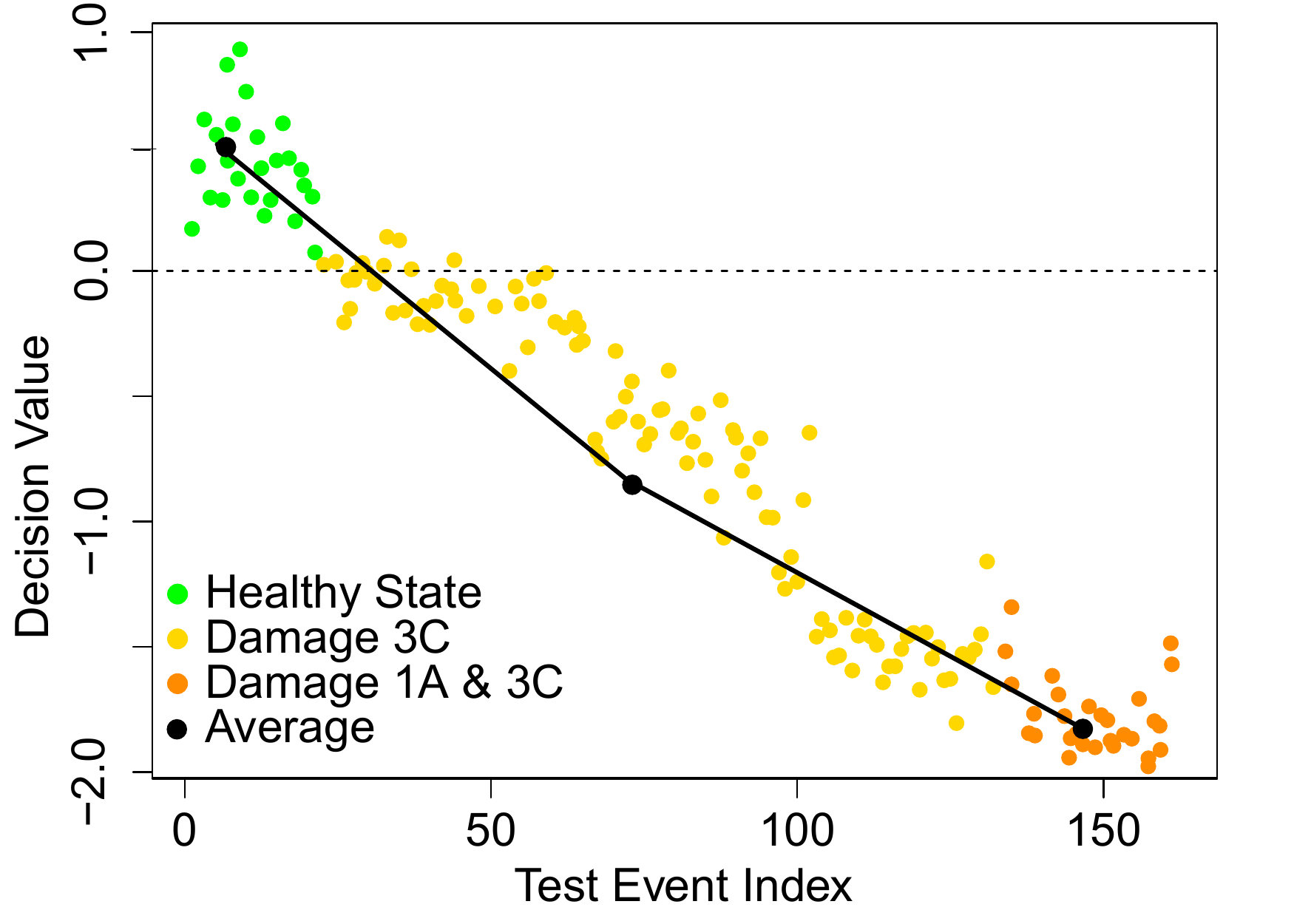} }}
	\subfloat[ SALS.] {{\includegraphics[height=1.5in,width=1.7in]{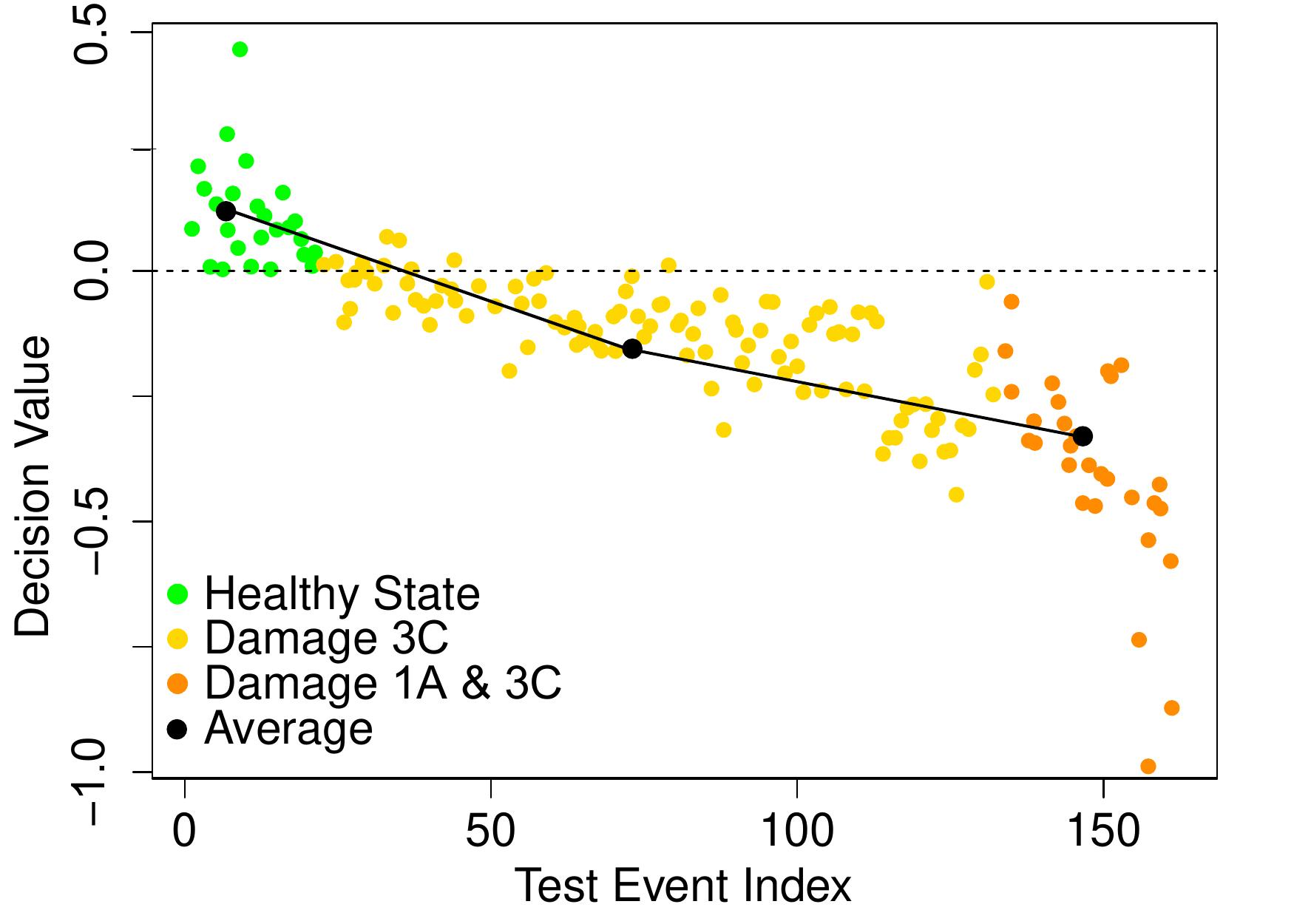} }}%
	\caption{Damage estimation applied on Building data using decision values obtained by  one-class SVM.}%
	\label{fig:dv_build}
\end{figure}

Similar to the Bridge dataset, we further analyzed the resultant decision values which were also able to characterize damage severity. Figure \ref{fig:dv_build} demonstrates that the more severe damage to the $1A$ and $3C$ location test data the more deviation from the training data with lower decision values. %Check the meaning%    

%that the more severe damage test data related to locations $1A$ and $3C$ were more deviated from the training data with lower  decision values.
\begin{figure}[!t]
	\captionsetup[subfloat]{justification=centering}
	\subfloat[NeCPD.] {{\includegraphics[height=2.2in,width=3.5in]{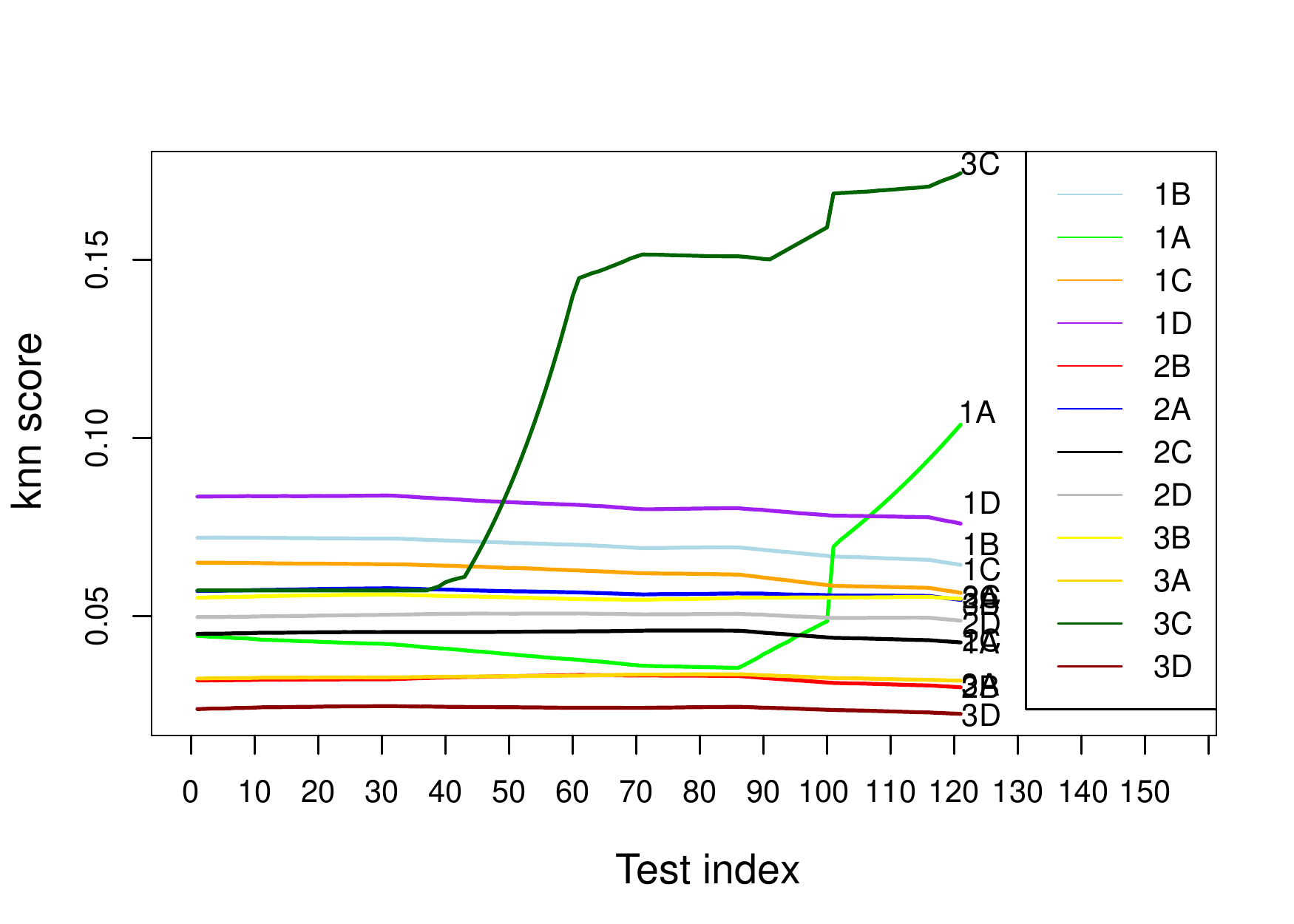} }}
	
	\subfloat[ SALS.] {{\includegraphics[height=2.2in,width=3.5in]{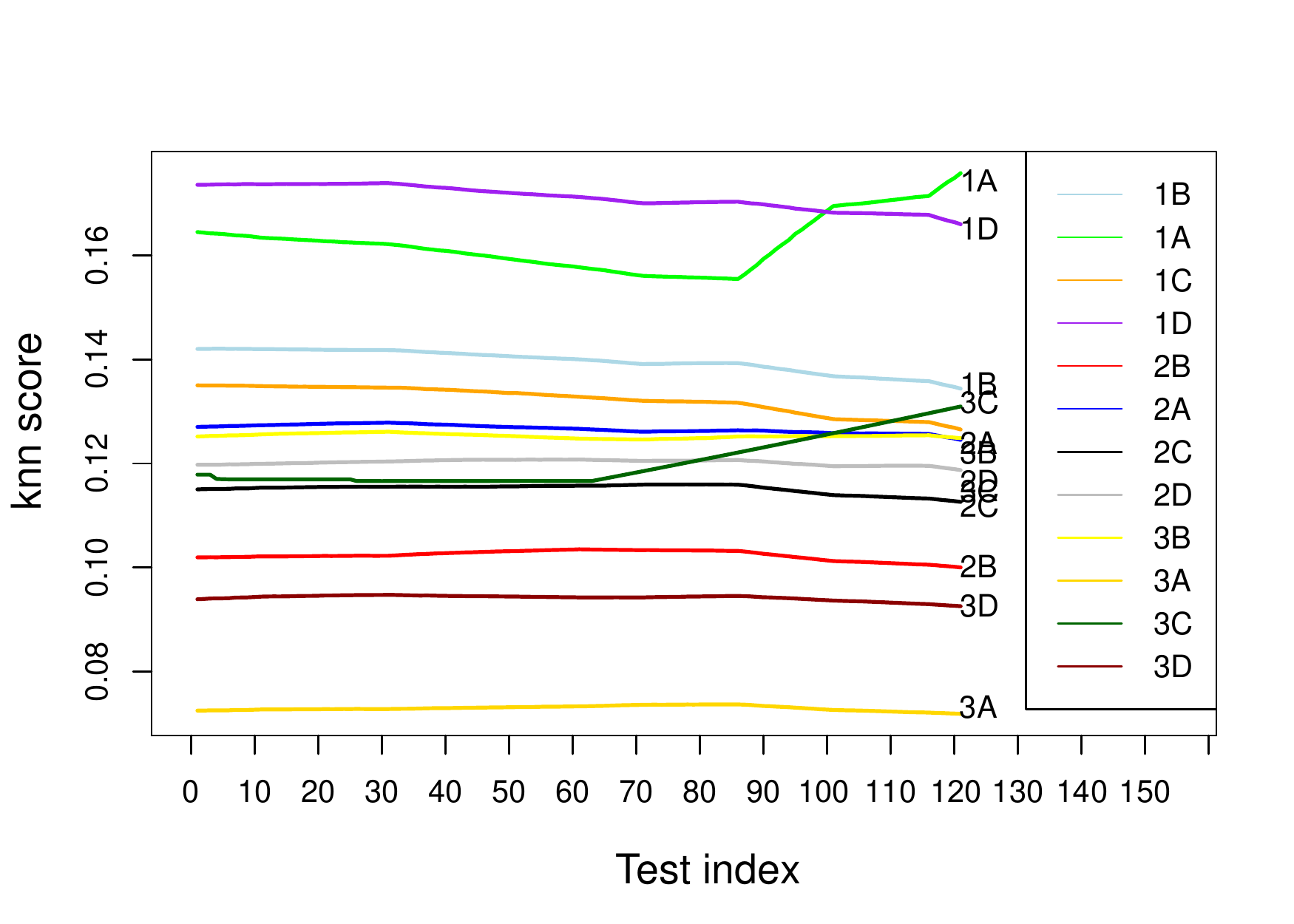} }}%
	
	%\subfloat[ batchCP-ALS.] {{\includegraphics[height=2.2in,width=3.5in]{build_batch_loc} }}%
	\caption{ Damage localization for the Building data: NeCPD  successfully localized  damage locations.}
	\label{fig:build_necpd_loc}
\end{figure}

 The last experiment is to compute the $k$-nn score for each sensor based on the $k$-nearest neighboring of the average distance between each row of the matrix $B_{new}$. Figure \ref{fig:build_necpd_loc} shows the resultant $k$-nn score for each sensor. The first 30 events (depicted on the x-axis) represent the healthy data, followed by 60 events describing when the damage was introduced in location $3C$. The last 30 events represent the damage occurred in both locations $1A$ and $3C$. It can be clearly observed that the NeCPD  method successfully localized the structure's damage where sensors $1A$ and $3C$ behave significantly different from all the other sensors apart from the position of the introduced damage. However, the SALS method is not able to successfully locate damage since it fails to update the location matrix $B$  incrementally.

\section{Conclusion}
\label{s:conclusion}

%In this paper, we introduce our new method NeCPD for tensor decomposition. The method derived based on well-known CP decomposition with stochastic gradient descent algorithm for multi-way analysis. 

This paper investigated the CP decomposition with stochastic gradient  descent  algorithm for multi-way data analysis. This leads to a new method named NeCPD for tensor decomposition. Our new method guarantees the convergence for a given non-convex problem by modeling the second order derivative of the loss function and incorporating little noise to the gradient update. Furthermore, NeCPD employs Nesterov's method to compensate for the delays of the optimization process and accelerate the convergence rate. Based on laboratory and real datasets from the area of SHM, our NeCPD, with one-class SVM model for anomaly detection, achieve accurate results in damage detection, localization, and assessment in online and one-class settings. Among the key future work is how to parallelize the tensor decomposition with NeCPD. Also, it would be useful to apply NeCPD with datasets from different domains.

%This paper investigated the CP decomposition with stochastic gradient  descent  algorithm for multi-way data analysis. This leads to a new algorithm named NeCPD for tensor decomposition which guarantees  the convergence for a given non-convex problem    by computing  the second order derivative of the loss  function and adding  little noise to the gradient update. It further applies Nesterov's   method   to compensate the delay of the optimization process and  accelerate the  convergence rate. This  new method with one-class SVM model for anomaly detection,  achieved promising results in damage detection, localization, and assessment in an online and one-class manner using laboratory-based and real-life structural  datasets in the area of SHM. Our future works will focus on how to parallelize the tensor decomposition with  NeCPD.

\section{Acknowledgement}
The authors wish to thank the Roads and Maritime Services (RMS) in New South Wales, New South Wales Government in Australia and  Data61
(CSIRO) for provision of the support and testing facilities for this research
work. Thanks are also extended to Western Sydney University for facilitating the experiments
on the cable-stayed bridge.

\bibliographystyle{IEEEtran}
\bibliography{myBib}

\end{document}